\documentclass[letterpaper, 10 pt, conference]{ieeeconf}  

\IEEEoverridecommandlockouts
\usepackage{cite}
\usepackage{amsmath,amssymb}
\usepackage{graphicx}
\usepackage{booktabs}
\usepackage{multirow}
\usepackage{algorithm}
\usepackage{algpseudocode}
\usepackage{xcolor}
\usepackage{hyperref}
\usepackage{subcaption}

\title{\LARGE \bf GraspADMM: Improving Dexterous Grasp Synthesis\\
via ADMM Optimization}

\author{Liangwang Ruan$^{1,2*}$, Jiayi Chen$^{1,2*}$, He Wang$^{1,2\dagger}$, Baoquan Chen$^{3\dagger}$%
\thanks{$^{1}$School of Computer Science, Peking University. $^{2}$Galbot.}%
\thanks{$^{3}$School of Intelligence Science and Technology, Peking University.}%
\thanks{$^*$Equal contribution. $^\dagger$Corresponding authors.}%
}

\newcommand{\lw}[1]{#1}
\algnewcommand{\LineComment}[1]{\Statex \textit{// #1}}

\begin{document}

\maketitle
\thispagestyle{empty}
\pagestyle{empty}

\begin{abstract}
Synthesizing high-quality dexterous grasps is a fundamental challenge in robot manipulation, \lw{requiring diversity, kinematic feasibility (valid hand-object contact without penetration), and dynamic stability (secure multi-contact forces)}. The recent framework Dexonomy successfully ensures broad grasp diversity through dense sampling and improves kinematic feasibility via a simulator-based refinement method that excels at resolving exact collisions. However, its reliance on fixed contact points restricts the hand's reachability and prevents the optimization of grasp metrics for dynamic stability. Conversely, purely gradient-based optimizers can maximize dynamic stability but rely on simplified contact approximations that inevitably cause physical penetrations. \lw{Building upon Dexonomy's sampling-based taxonomy pipeline}, we propose GraspADMM, an ADMM-based refinement framework that replaces Dexonomy's fixed-contact simulator refinement with decoupled optimization over object-side target contacts and hand-side realized contacts. This decomposition allows the pipeline to alternate between updating the target object points to directly maximize dynamic grasp metrics, and adjusting the hand pose to physically reach these targets while strictly respecting collision boundaries. Simulation experiments demonstrate that GraspADMM significantly outperforms state-of-the-art baselines, achieving a nearly 15\% absolute improvement in grasp success rate for type-unaware synthesis and roughly a 100\% relative improvement in type-aware synthesis. Furthermore, our approach maintains robust, physically plausible grasp generation even under extreme low-friction conditions.
\end{abstract}

\section{Introduction}

Dexterous grasping is essential for general-purpose robot manipulation. With high degrees of freedom (DoF, typically $>20$), dexterous hands can adapt to diverse object shapes. However, this flexibility makes grasp planning highly complex. While data-driven methods have succeeded for parallel grippers by scaling up dataset sizes~\cite{fang2020graspnet,Fang2023AnyGrasp}, their extension to dexterous hands is limited by a lack of high-quality datasets. Building these datasets requires solving four fundamental challenges: (1) \textit{diversity}, ensuring broad coverage of objects, hand poses, and grasp types; (2) \textit{kinematic feasibility}, achieving hand-object contact without penetration; (3) \textit{dynamic stability}, generating sufficient contact forces to balance the object's gravity; and (4) \textit{speed}, achieving fast synthesis. 

To address these requirements, analytic methods that optimize metrics like force closure~\cite{ferrari1992planning} have been explored. Frameworks like GraspIt!~\cite{miller2004graspit} used sampling to handle these non-differentiable metrics, but were highly inefficient for high-DoF hands. Recent works~\cite{liu2021synthesizing, turpin2022grasp, turpin2023fast, wang2023dexgraspnet, chen2023task, li2023frogger, chen2024springgrasp, chen2024bodex} improved efficiency by introducing differentiable grasp metrics for gradient-based optimization. However, they often focus only on specific grasp types (e.g., fingertips) and suffer deep penetrations, leading to poor diversity and kinematic feasibility.

Recently, Dexonomy~\cite{chen2025dexonomy} proposed a hybrid approach that 	improves upon prior work. It ensures broad grasp diversity through human-annotated templates and dense sampling, and achieves competitive synthesis speeds. Notably, Dexonomy introduces simulator-based refinement for kinematic feasibility. By utilizing transposed Jacobian control—acting as virtual springs—to pull the hand toward target object points within the forward step of a physics simulator, it robustly resolves exact collisions, demonstrating much better kinematic feasibility than previous gradient-based works.

\begin{figure}[t]
    \centering
    \includegraphics[width=0.9\linewidth]{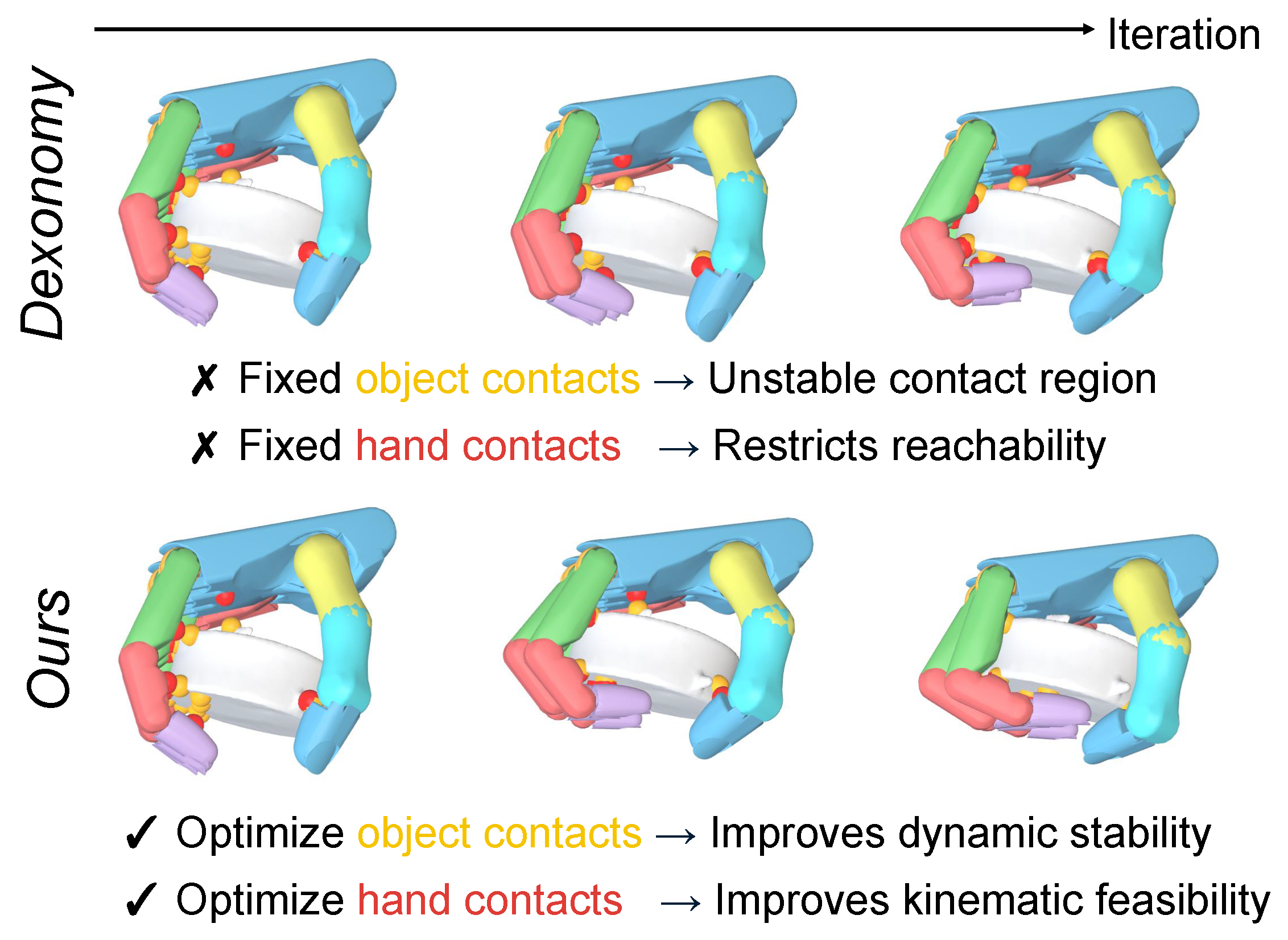}
    \caption{\textbf{Overview}.
    From the same initialization, our GraspADMM framework generates robust grasps by optimizing contact points on the hand (red) and the object (yellow), while Dexonomy~\cite{chen2025dexonomy} simply fixes these points.}
    \label{fig:teaser}
    \vspace{-3mm}
\end{figure}

\begin{table*}[t]
\centering
\begin{tabular}{l c c c c}
\toprule
\textbf{Methodology} & \textbf{Diversity} & \textbf{Kinematic Feasibility} & \textbf{Dynamic Stability} & \textbf{Speed}\\
\midrule
Sampling-based (e.g., GraspIt!~\cite{miller2004graspit}) & $\times$ & $\times$ & $\sim$ & $\times$\\
Gradient-based (e.g., MALA~\cite{liu2021synthesizing}) & $\sim$ & $\times$ & $\checkmark$ & $\sim$ \\
Simulator-based (Dexonomy~\cite{chen2025dexonomy}) & $\checkmark$ & $\sim$ & $\sim$ & \textbf{$\checkmark$} \\
\midrule
\textbf{GraspADMM (Ours)} & \textbf{$\checkmark$} & \textbf{$\checkmark$} & \textbf{$\checkmark$} & \textbf{$\checkmark$} \\
\bottomrule
\end{tabular}%
\label{tab:method_comparison}
\caption{\textbf{Comparison of Dexterous Grasp Synthesis Paradigms.} We evaluate existing methodologies based on their ability to satisfy the four core requirements of high-quality grasp generation. ($\checkmark$: Fully supports; $\sim$: Partially supports / Some work supports; $\times$: Struggles).}
\end{table*}

However, Dexonomy does not fully exploit the potential of this simulator-based framework, leaving critical room for improvement. Specifically, its refinement stage fixes both the target object points and the assigned hand contact points (Fig.~\ref{fig:teaser}), which introduces two fundamental drawbacks:
\begin{enumerate}
    \item \textbf{Fixed object points limit dynamic stability:} The object contact points may be established in suboptimal regions, where the resultant contact force cannot hold the object. Although Dexonomy uses grasp metrics as a post-hoc filter, it simply discards unstable grasps and cannot actively improve them.
    \item \textbf{Fixed hand points limit kinematic feasibility:} kinematic constraints may prevent hand points from reaching their targets simultaneously.
\end{enumerate}

Overcoming these drawbacks to unify strict collision resolution with active dynamic optimization is highly non-trivial. On one hand, off-the-shelf physics simulators excel at exact collision resolution (kinematics), but they do not natively support incorporating grasp metrics as optimization objectives. On the other hand, purely gradient-based optimizers can directly maximize dynamic grasp metrics, but they struggle to prevent collisions because they often rely on simplified, penetration-prone contact approximations to bypass non-differentiable physics.

In this work, we focus solely on improving Dexonomy's refinement stage. GraspADMM inherits its sampling pipeline and simulator-based collision handling, but replaces fixed-contact local refinement with an ADMM formulation that introduces object- and hand-side contact mobility.

To achieve this, our formulation explicitly separates the optimization variables. We treat the target contact points on the object surface and the actual contact locations on the hand as independent variables, updating them alternately. This separation circumvents the need to backpropagate gradients through complex, discontinuous contact physics, enabling the use of exact mesh-to-mesh collision detection to rigorously maintain kinematic feasibility. Furthermore, this decomposition allows us to independently optimize the target object points to minimize grasp metrics, actively driving dynamic stability. Consequently, this decoupled approach efficiently synthesizes high-quality grasps that strictly satisfy both kinematic and dynamic constraints.

Simulation experiments demonstrate that GraspADMM significantly outperforms state-of-the-art baselines. On a large-scale benchmark of over 5,600 objects, our method achieves a nearly 15\% absolute improvement in grasp success rate over Dexonomy~\cite{chen2025dexonomy} for type-unaware synthesis, while maintaining zero penetration depth and similar speed. This performance gap widens in type-aware tasks, where our decoupled ADMM formulation delivers a 20\% to 30\% absolute increase (roughly a 100\% relative improvement) in success rates across diverse grasp types. Furthermore, under extreme low-friction conditions ($\mu=0.1$) where prior methods catastrophically fail, GraspADMM still consistently generates diverse and highly stable multi-fingered grasps.

\noindent\textbf{Our main contributions are:} 1) an ADMM-based contact refinement formulation that decouples object target contacts from hand realized contacts; 2) a practical alternating solver that combines differentiable grasp-quality optimization with MuJoCo-based penetration-free hand updates; and 3) empirical evidence that this refinement improves over Dexonomy across type-unaware, type-aware, and low-friction settings.

\section{Related Work}

\subsection{Analytical Grasp Synthesis}
In analytical grasp synthesis, algorithms typically assume complete geometric knowledge of an object to generate hand poses, which are subsequently evaluated using analytical grasp quality metrics. The most widely adopted metrics are based on force closure~\cite{ferrari1992planning, roa2015grasp}; however, these are traditionally non-differentiable. Consequently, early approaches either sampled a massive volume of grasps to select the highest-scoring candidates~\cite{mahler2016dex, mahler2018dex, fang2020graspnet, cao2021suctionnet} or employed sampling-based optimization techniques such as simulated annealing~\cite{miller2004graspit, ciocarlie2007dexterous}. While effective for simple parallel grippers and suction cups with low degrees of freedom (DoF $<8$), these strategies suffer from the curse of dimensionality and become computationally intractable for dexterous hands, where the DoF often exceeds $20$.

To overcome this limitation, recent research has shifted toward differentiable grasp metrics coupled with gradient-based optimization~\cite{liu2021synthesizing, turpin2022grasp, turpin2023fast, wang2023dexgraspnet, chen2023task, li2023frogger, chen2024springgrasp, chen2024bodex}. These works primarily focus on formulating novel metrics, such as those based on differentiable force closure~\cite{liu2021synthesizing,wang2023dexgraspnet} or differentiable physics simulations~\cite{turpin2022grasp, turpin2023fast}. Regarding optimization strategies, the Metropolis-adjusted Langevin Algorithm (MALA) has been utilized to enhance grasp diversity~\cite{liu2021synthesizing, wang2023dexgraspnet}. Furthermore, Dexonomy~\cite{chen2025dexonomy} highlights the critical role of initialization, combining taxonomy-guided sampling with simulator-based local refinement. Our framework differs from both prior gradient-based methods and Dexonomy: ADMM separates grasp-metric optimization from simulator-enforced contact feasibility, avoiding approximate contact models without fixing contact points.

An alternative line of research explores contact-region mapping, transferring human grasp demonstrations to novel objects via functional correspondence~\cite{yang2022oakink, wei2024learning, wu2024cross}. Although this produces semantically meaningful, human-like grasps suitable for high-level task planning, such approaches often sacrifice grasp diversity and struggle to guarantee dynamic stability under varying physical conditions.


\subsection{Alternating Direction Method of Multipliers (ADMM)}
The Alternating Direction Method of Multipliers (ADMM) is a well-established technique in convex optimization~\cite{Boyd2004convex}, \lw{proving particularly effective} for constrained and non-smooth optimization problems. By introducing auxiliary variables, ADMM decouples complex, nonlinear optimization problems into multiple simpler sub-problems, significantly accelerating convergence. This property has been widely exploited in computer graphics and physics simulation, such as in Projective Dynamics (PD)~\cite{Liu14mass,Narain16admm} and progressive mesh parameterization~\cite{Liu2018pp}. 

ADMM is also highly adept at handling complex constraints, successfully simulating massive non-penetration scenarios like collisions between elastic bodies and hair~\cite{Daviet2020simple,Daviet2023interactive}. Furthermore, by leveraging proximal operators, ADMM offers unique advantages for non-smooth objectives across geometry processing~\cite{Bouaziz2013sparse}, computer animation~\cite{Neumann2013sparse}, and image processing~\cite{Hu13fast}. Beyond applied domains, significant theoretical work has focused on improving the ADMM algorithm itself, including Anderson Acceleration~\cite{Zhang2019accelerating}, stochastic ADMM formulations~\cite{Bian2021stocastic}, and extensions to non-convex optimization~\cite{wang2018globalconvergenceadmmnonconvex}. Inspired by its success in resolving discontinuous physical constraints in simulation, our work adapts ADMM for dexterous grasp synthesis, utilizing its decoupled nature to satisfy non-smooth exact collision boundaries while simultaneously optimizing dynamic stability.
\section{Method}

\subsection{Pipeline Overview}

Our grasp synthesis framework builds upon the foundational pipeline of Dexonomy~\cite{chen2025dexonomy}, progressing through four primary stages: template annotation, sampling-based initialization, simulator-based refinement, and template bootstrapping. All non-refinement stages follow Dexonomy; the core technical innovation of GraspADMM lies in replacing Dexonomy's fixed-contact kinematic refinement with the ADMM formulation in Sec.~\ref{sec:admm}, which jointly optimizes grasp metrics and penetration-free contact feasibility.

To contextualize this contribution, we first briefly review the preliminary initialization stages inherited from the Dexonomy architecture. The pipeline begins with human-annotated grasp templates mapped to specific categories within the GRASP taxonomy~\cite{feix2015grasp}. These templates include the hand pose $\mathbf{q}\in \mathbb{R}^q$, hand contact points $\mathbf{p}_i^h\in \mathbb{R}^3$, corresponding inward-pointing normals $\mathbf{n}_i^h\in \mathbb{R}^3$, and the link assignments for each contact point, where $i \in \{1, 2, \dots, m\}$. 

During the initialization stage, the method establishes an initial contact alignment between the object and the grasp template through a hybrid strategy of dense sampling and massively parallel GPU optimization. Specifically, the system randomly selects a grasp template, a hand contact point, and an object surface point. It initializes the object pose by matching the contact locations while aligning the contact normals in opposing directions. Subsequently, with the hand pose fixed, the method computes the nearest surface projections for all hand contact points and optimizes the object's rigid transformation by minimizing the geometric discrepancy between the hand and object contact points. After this optimization, the results undergo sequential post-filtering to remove severe physical penetrations.

In the ADMM optimization stage, we fix the object pose and optimize the hand pose $\mathbf{q}$, along with auxiliary contact variables, to minimize the dynamic grasp quality metric. Upon convergence, we filter out suboptimal solutions that exhibit poor grasp metrics, unresolved geometric penetrations, or a failure to meet the topological contact requirements defined by the template. Successful grasps are then passed to a rigorous physics simulator evaluation stage and subsequently used to update the templates. \lw{GraspADMM is a refinement module rather than a replacement for the entire Dexonomy pipeline, it can in principle be paired with other initialization methods that provide feasible or near-feasible contact templates.} For comprehensive details regarding the non-optimization components of this pipeline, please refer to Dexonomy~\cite{chen2025dexonomy}.

\subsection{Grasp Quality Metric}

While Dexonomy only uses \lw{the grasp quality metric as a post-hoc filter, our work uses it as} the optimization objective and thus greatly improves dynamic stability. In this section, we summarize the force-closure grasp quality metric utilized in this paper, which aligns with prior  works~\cite{chen2024bodex,chen2025dexonomy}. Note that our algorithm is agnostic to the specific metric and can accommodate other differentiable grasp metrics. For each contact $i$, let $\mathbf{p}_i \in \mathbb{R}^3$ denote the contact position, $\mathbf{n}_i \in \mathbb{R}^3$ the inward-pointing surface unit normal on the object, and $\mathbf{d}_i \in \mathbb{R}^3$ and $\mathbf{c}_i \in \mathbb{R}^3$ be two orthogonal unit tangent vectors satisfying $\mathbf{n}_i = \mathbf{d}_i \times \mathbf{c}_i$. The Coulomb friction cone $\mathcal{F}_i$ and the contact Jacobian $\mathbf{J}_{o,i}$ for the object at contact $i$ are defined as follows:
\begin{align}
\label{eq: F_pcf}
\mathcal{F}_i & = \left\{\mathbf{x}_i\in\mathbb{R}^3~|~0 \leq x_{i,1} \leq 1, x_{i,2}^2+x_{i,3}^2 \leq \mu^2 x_{i,1}^2 \right\}, \\
\label{eq: G_pcf}
\mathbf{J}_{o,i}^T & = 
\begin{bmatrix}
    \mathbf{n}_i & \mathbf{d}_i & \mathbf{c}_i \\
    \mathbf{p}_i \times \mathbf{n}_i & 
    \mathbf{p}_i \times \mathbf{d}_i & 
    \mathbf{p}_i \times \mathbf{c}_i \\
\end{bmatrix} \in \mathbb{R}^{6\times3},
\end{align}
where $\mu$ is the friction coefficient. The friction cone $\mathcal{F}_i$ represents all physically feasible contact forces at contact $i$, and $\mathbf{J}_{o,i}$ maps a local contact force $\mathbf{x}_i$ to a global object wrench $\mathbf{w}_i=\mathbf{J}_{o,i}^T\mathbf{x}_{i}$.

To balance an external wrench $\mathbf{g}\in\mathbb{R}^6$ (e.g., object gravity), the optimal contact forces $\{\mathbf{f}_i\}^m_{i=1}$ are obtained by solving the following quadratic program (QP):
\begin{align}
  (\mathbf{f}_1, \dots, \mathbf{f}_m) = \underset{(\mathbf{x}_1, \dots, \mathbf{x}_m)}{\arg \min}~~~&\|\sum_{i=1}^m\mathbf{J}_{o,i}^T\mathbf{x}_{i} - \mathbf{g}\|^2 \label{eq: qp force}, \\
    \text{s.t.}~~~&\mathbf{x}_{i} \in \mathcal{F}_i, ~~i\in\{1,\dots,m\}, \\
    &\sum_{i=1}^m x_{i, 1} \geq \lambda_{\text{min}}, \label{eq: normal force constraint}
\end{align}
where $\lambda_{\text{min}}$ is a hyperparameter enforcing a minimum total normal force to prevent the solver from returning trivial zero-force solutions. To reduce computational complexity, the nonlinear friction cone $\mathcal{F}_i$ is approximated by a polyhedral pyramid, converting the problem into a linearly-constrained QP that can be efficiently solved using Clarabel~\cite{goulart2024clarabel}.

Finally, the dynamic grasp quality metric $e$ is defined as the residual wrench magnitude: 
\begin{equation}
e=\|\sum_{i=1}^m\mathbf{J}_{o,i}^T\mathbf{f}_{i} - \mathbf{g}\|^2, \label{eq: qp quality}
\end{equation}
where a lower $e$ indicates a more stable and robust grasp. Following~\cite{wu2022learning}, we set $\mathbf{g} = \mathbf{0}$ in Eq.~\ref{eq: qp force} rather than systematically testing six orthogonal gravity vectors as done in~\cite{chen2024bodex}, \lw{while the final success criterion is still evaluated by MuJoCo rollouts under six external perturbation forces. Thus this approximation only guides contact refinement and does not replace physical validation}

\subsection{ADMM Optimization}
\label{sec:admm}

Starting from the initialized pose $\mathbf{q}$, the optimization problem in this stage is formally defined as:
\begin{equation}
    \min_{\mathbf{q} \in \mathcal{C}\bigcap \mathcal{T}} e(\mathbf{p}^o, \mathbf{n}^o).
    \label{eq:local-opt}
\end{equation}
Here, $e$ is the QP grasp quality metric from Eq.~\ref{eq: qp quality}, formulated as a function of the target object contact points $\mathbf{p}^o=\{\mathbf{p}_i^o\}$ and their corresponding normals $\mathbf{n}^o=\{\mathbf{n}_i^o\}$. The constraint $\mathcal{C}$ denotes the strictly penetration-free kinematic workspace of the hand, and $\mathcal{T}$ represents the hand-object contact constraint enforcing $\mathbf{p}_i^h=\mathbf{p}_i^o$ (i.e., the hand contact points must exactly touch the target object points). 

Instead of solving this highly nonlinear constrained optimization directly, we employ ADMM to decouple the hard contact constraint $\mathcal{T}$. This yields the following augmented Lagrangian:
\begin{equation}
\begin{aligned}
    \mathcal{L}(\mathbf{p}^o, \mathbf{p}^h, \lambda) &= e(\mathbf{p}^o, \mathbf{n}^o)   + \rho \lambda^T ( \mathbf{p}^h - \mathbf{p}^o) + \frac{\rho}{2} \lVert \mathbf{p}^h - \mathbf{p}^o \rVert^2 \\
    &= e(\mathbf{p}^o, \mathbf{n}^o) + \frac{\rho}{2} \lVert \mathbf{p}^h - \mathbf{p}^o + \lambda \rVert^2 - \frac{\rho}{2} \lVert \lambda \rVert^2,
\end{aligned}
\label{eq:local-langrangian}
\end{equation}
where $\lambda$ is the scaled dual variable (Lagrange multiplier) and $\rho$ is the penalty stiffness parameter. The original problem in Eq.~\ref{eq:local-opt} is thus reformulated into a saddle-point problem:
\begin{equation}
    \min_{\mathbf{q}\in \mathcal{C}, \mathbf{p}^h, \mathbf{p}^o} \max_{\lambda} \mathcal{L}(\mathbf{p}^o, \mathbf{p}^h, \lambda).
    \label{eq:local-saddle}
\end{equation}
We solve Eq.~\ref{eq:local-saddle} through the following three-step alternating minimization process:

\begin{figure}[t]
    \centering
    \includegraphics[width=1\linewidth]{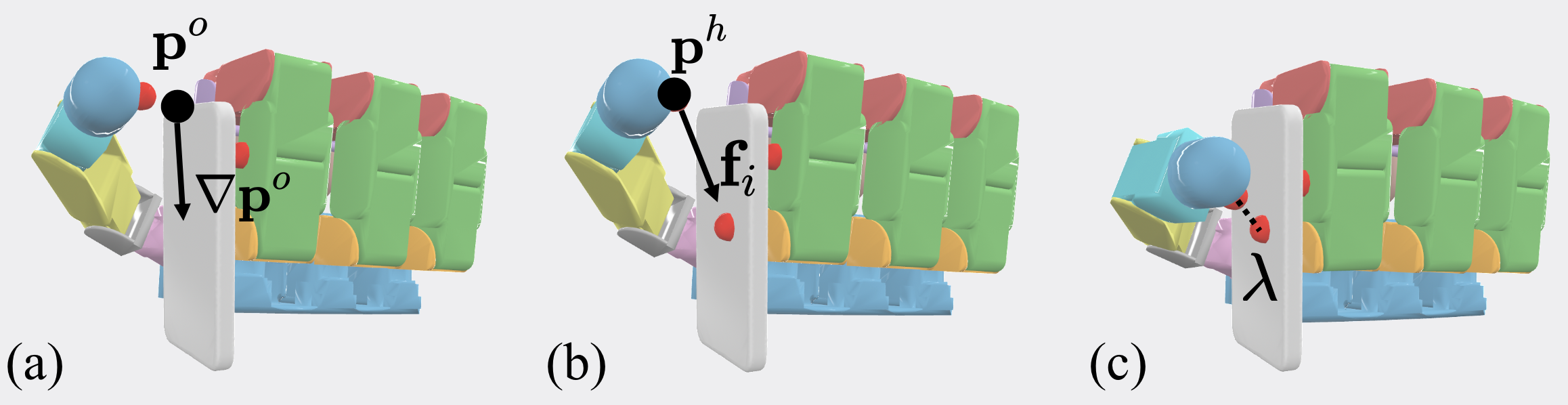}
    \caption{\textbf{ADMM Optimization Pipeline}. (a) Update target object contact points $\mathbf{p}^o$ via gradient descent to maximize dynamic stability. (b) Update hand pose $\mathbf{q}$ and points $\mathbf{p}^h$ via forward-simulated transposed Jacobian control to satisfy kinematic feasibility. (c) Update the dual variable $\lambda$.}
    \label{fig:admm-opt}
\end{figure}

\paragraph{1. Fix $\mathbf{q},\mathbf{p}^h,\lambda$, update $\mathbf{p}^o$} The subproblem for the target object points is:
\begin{equation}
    \min_{\mathbf{p}^o}  e(\mathbf{p}^o, \mathbf{n}^o) + \frac{\rho}{2} \lVert \mathbf{p}^h - \mathbf{p}^o + \lambda \rVert^2,
    \label{eq:admm-1}
\end{equation}
where $\mathbf{p}^o$ is strictly constrained to the object surface. We use a one-step gradient descent to optimize Eq.~\ref{eq:admm-1} and project $\mathbf{p}^o$ back onto the object surface after each update. To compute the gradient of the surface normal $\mathbf{n}^o$ with respect to $\mathbf{p}^o$, we utilize barycentric interpolation across the local mesh triangles, following the methodology in~\cite{turpin2023fast}.

\paragraph{2. Fix $\mathbf{p}^o,\lambda$, update $\mathbf{q},\mathbf{p}^h$} The subproblem for the hand kinematics is:
\begin{equation}
    \min_{\mathbf{q}\in \mathcal{C},\mathbf{p}^h} \frac{\rho}{2} \lVert \mathbf{p}^h - \mathbf{p}^o + \lambda \rVert^2.
\end{equation}
This step is mathematically equivalent to minimizing the Euclidean distance between the hand points $\mathbf{p}^h$ and the shifted target points $(\mathbf{p}^o - \lambda)$ while strictly maintaining an interpenetration-free hand state ($\mathbf{q}\in \mathcal{C}$). To robustly optimize this non-convex spatial target, we utilize forward physics simulation in MuJoCo. At each hand contact point $\mathbf{p}^h_i$, we apply a virtual spring force $\mathbf{f}_i$ to bridge the distance gap. This force is mapped to the hand's joint torques $\mathbf{\tau}$ via transposed Jacobian control:
\begin{equation}
    \mathbf{f}_i = k_f (\mathbf{p}^o - \lambda - \mathbf{p}^h), ~~~ \mathbf{\tau} = \sum_{i=1}^m \mathbf{J}_{h,i}^T \mathbf{f}_i, 
    \label{eq:local-force}
\end{equation}
where $k_f$ is the virtual spring stiffness, and $\mathbf{J}^T_{h,i}\in \mathbb{R}^{q\times 3}$ is the transpose of the hand's contact Jacobian. By stepping the simulation forward, MuJoCo automatically and robustly resolves all physical collisions. Since we allow $\mathbf{p}_i^h$ to slide freely within its assigned hand link, we project the target point $\mathbf{p}_i^o$ onto the surface of the corresponding hand link to establish the updated $\mathbf{p}_i^h$ at each timestep.

\begin{algorithm}[t]
\caption{GraspADMM Optimization Pipeline}
\label{alg:admm}
\begin{algorithmic}[1]
\Require Initial pose $\mathbf{q}_0$ (from Dexonomy initialization); Object mesh $\mathcal{M}$; Template $\mathcal{P}$; Penalty $\rho$
\Ensure Optimized hand pose $\mathbf{q}^*$

\LineComment{\textit{Initialize from Dexonomy prior}}
\State $\mathbf{q}, \lambda  \gets \mathbf{q}_0, 0$ 
\State $\mathbf{p}^o, \mathbf{p}^h \gets \text{InitializeContacts}(\mathbf{q}, \mathcal{P}, \mathcal{M})$

\While{not converged \textbf{and} iter $< K$}
    \LineComment{\textit{Step 1: Object Contact Update}}
    \State $\mathbf{p}^o \gets \mathbf{p}^o - \alpha \nabla_{\mathbf{p}^o} \mathcal{L}(\mathbf{p}^o, \mathbf{p}^h, \lambda)$ \Comment{Eq.~\ref{eq:admm-1}}
    \State $\mathbf{p}^o \gets \text{ProjectToSurface}(\mathbf{p}^o, \mathcal{M})$
    
    \Statex
    \LineComment{\textit{Step 2: Hand Kinematics Update via Simulation}}
    \State $\mathbf{f}_i \gets k_f (\mathbf{p}^o_i - \lambda_i - \mathbf{p}^h_i), \quad \forall i \in \{1 \dots m\}$ \Comment{Eq.~\ref{eq:local-force}}
    \State $\mathbf{\tau} \gets \sum_{i=1}^m \mathbf{J}_{h,i}^T \mathbf{f}_i$ \Comment{Transposed Jacobian control}
    \State $\mathbf{q}, \mathbf{p}^h \gets \text{StepMuJoCo}(\mathbf{q}, \mathbf{\tau})$ \Comment{Forward simulation}
    
    \Statex
    \LineComment{\textit{Step 3: Dual Variable Update \& Annealing}}
    \State $\lambda \gets \lambda + (\mathbf{p}^h - \mathbf{p}^o)$ \Comment{Eq.~\ref{eq:lam-update}}
    \State $\lambda \gets \text{Annealing}(\lambda)$ \Comment{Mitigate Divergence}
\EndWhile

\State \Return $\mathbf{q}$
\end{algorithmic}
\end{algorithm}

\paragraph{3. Fix $\mathbf{q},\mathbf{p}^h,\mathbf{p}^o$, update $\lambda$} We update the dual variable using the standard ADMM dual ascent step:
\begin{equation}
    \lambda \gets \lambda + \mathbf{p}^h - \mathbf{p}^o.
    \label{eq:lam-update}
\end{equation}

\lw{Though our formulation preserves the variable-splitting structure of ADMM, because the hand-update subproblem is solved by nonconvex MuJoCo simulation with hard contact constraints, we do not claim classical ADMM convergence guarantees. If the residual error $\mathbf{p}^h - \mathbf{p}^o$ accumulates across multiple steps, the dual variable $\lambda$ may diverge: the hand target point $\rm{p}^o - \lambda$ is hovering far in space rather than attached to the object.} To ensure stability, we implement a dual-variable annealing strategy: if the magnitude $\lVert \lambda_i \rVert > 2\times 10^{-2}$ for any contact point $i$ during the optimization loop, we manually reset $\lambda_i=0$ immediately after Eq.~\ref{eq:lam-update}.

The decoupled nature of our ADMM formulation can be intuitively understood by examining the extremes of the penalty parameter $\rho$. If $\rho=0$, the Lagrangian in Eq.~\ref{eq:local-langrangian} reduces entirely to the grasp metric $e$. In this scenario, the algorithm simply optimizes the target points on the object surface without considering the hand's kinematic reachability, inevitably resulting in dynamic targets the hand physically cannot grasp. Conversely, as $\rho \to \infty$ (or if we set $e=0$), the Lagrangian is dominated by the geometric constraint $\mathbf{p}^h = \mathbf{p}^o$. This reduces the system to a purely kinematic solver—much like the baseline Dexonomy refinement—where the hand perfectly matches the object surface but operates entirely blind to physical stability forces. By selecting an appropriate intermediate value ($\rho=10^3$ in our experiments), GraspADMM effectively balances these two forces, allowing the algorithm to dynamically hunt for stable force-closure configurations while rigorously respecting the kinematic boundaries of the initialized grasp pose.

Ultimately, by deploying ADMM, we fully decouple the optimization of the abstract grasp metric from the handling of hard collision constraints. This eliminates the need to backpropagate gradients through discontinuous, non-differentiable collision physics. Because each decoupled subproblem is significantly less nonlinear than the joint problem, they can be efficiently solved using specialized optimizers (gradient descent for $e$, MuJoCo for $\mathcal{C}$). Consequently, as demonstrated in Table~\ref{tab: fingertip baseline}, our framework achieves significantly faster convergence and dramatically fewer penetrations compared to traditional end-to-end optimization methods.
\section{Experiments}

\subsection{Evaluation Metric}
\label{sec:metric-caption}
Our experiment settings and evaluation metrics follow Dexonomy~\cite{chen2025dexonomy}. We make a brief summary below:
\begin{itemize}
    \item \textbf{GSR (\%):} Grasp Success Rate, the percentage of successful grasps out of total input attempts. A grasp succeeds only if it resists six external forces in MuJoCo. The success criteria for the object pose are 5cm and 15$^\circ$.
    \item \textbf{OSR (\%):} Object Success Rate, the percentage of objects for which at least one successful grasp is found.
    \item \textbf{CDC (mm):} Contact Distance Consistency, the delta between the maximum and minimum signed distances across all fingers. A smaller CDC indicates uniform, high-quality contact across all engaged fingers.
    \item \textbf{CLN:} Contact Link Number, the number of hand links whose distance to the object surface is within 2 mm\lw{.}
    \item \textbf{PD (mm):} Penetration Depth, encompassing both hand-object inter-penetrations and hand self-collisions.
    \item \textbf{Div (\%):} Diversity, measured by the proportion of the first eigenvalue to the sum of all eigenvalues in the Principal Component Analysis (PCA) of the hand poses.
\end{itemize}

\begin{table}[t]
    \centering
    \begin{tabular}{l c c c c c}
        \toprule
        \textbf{Method} & \textbf{GSR}$\uparrow$ & \textbf{OSR}$\uparrow$ & \textbf{CDC}$\downarrow$ & \textbf{PD}$\downarrow$ & \textbf{Div}$\downarrow$ \\
        \midrule
        DexGraspNet~\cite{wang2023dexgraspnet} & 12.1 & 57.0 & 7.6 & 4.9 & 29.0 \\
        FRoGGeR~\cite{li2023frogger} & 10.3 & 55.7 & 5.0 & 0.2 & 27.0 \\
        SpringGrasp~\cite{chen2024springgrasp} & 7.8 & 35.4 & 23.6 & 16.6 & 70.2 \\
        BODex~\cite{chen2024bodex} & 49.2 & 96.6 & 3.0 & 0.6 & 32.5 \\
        Dexonomy~\cite{chen2025dexonomy} & 60.5 & 96.5 & 0.21 & \textbf{0.0} & 34.2 \\
        \midrule
        \textbf{GraspADMM (Ours)} & \textbf{74.6} & \textbf{97.2} & \textbf{0.17} & \textbf{0.0} & \textbf{25.7} \\
        \bottomrule
    \end{tabular}%
    \caption{\textbf{Type-Unaware Grasp Synthesis Comparison.} Our method greatly outperforms Dexonomy on grasp success rate and diversity.}
    \label{tab: fingertip baseline}
\end{table}

\begin{figure}
    \centering
    \includegraphics[width=1\linewidth]{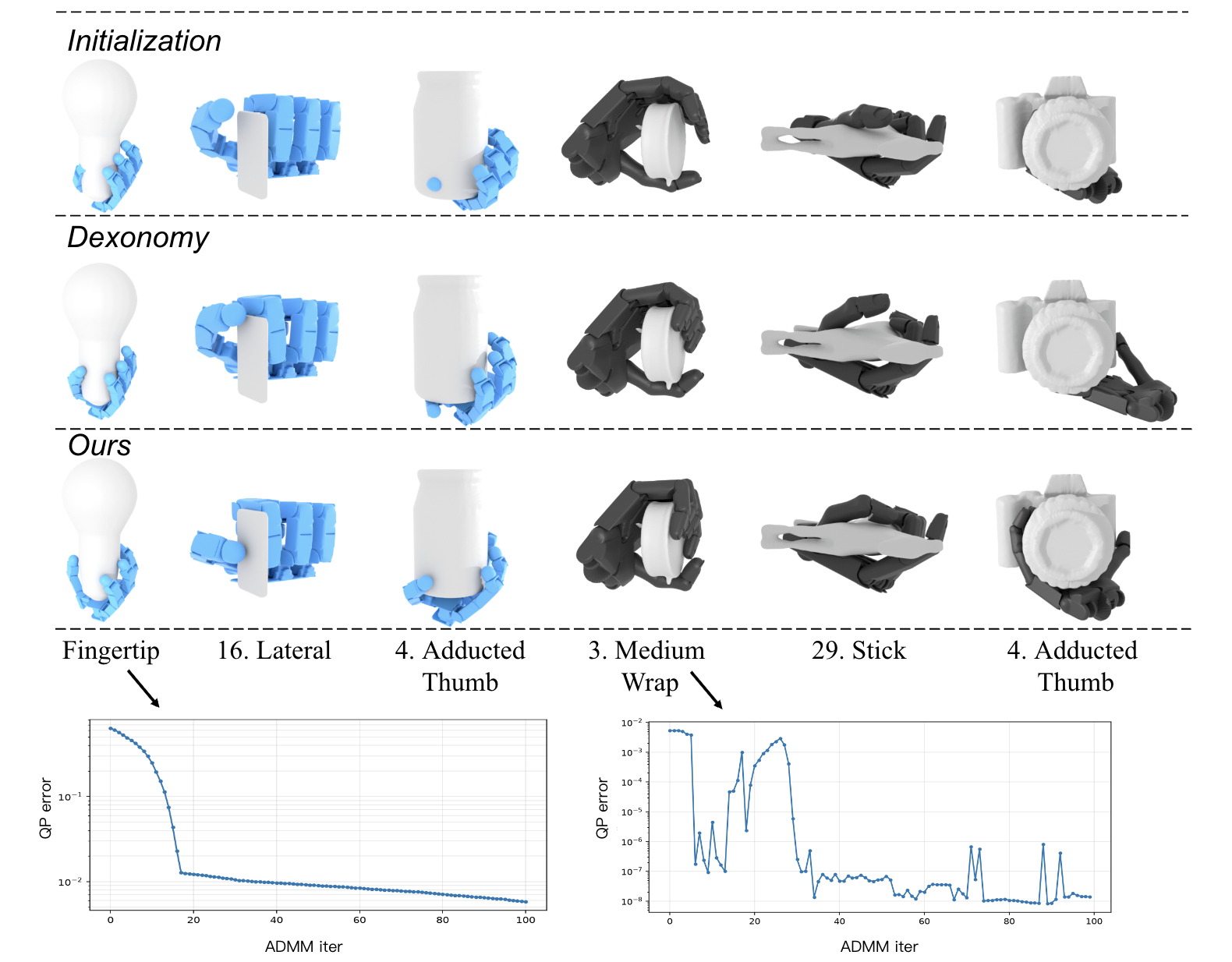}
    \caption{\textbf{Visualization of Optimized Grasps.} Comparison using the Allegro hand (blue) and Shadow hand (black). Given identical initializations, our method synthesizes more physically stable grasps than Dexonomy. \lw{Two representative ADMM convergence curves are also shown here.}}
    \label{fig:vis_compare}
\end{figure}

\subsection{Type-unaware Grasp Synthesis}

We adopt the identical experimental setting as Dexonomy~\cite{chen2025dexonomy}: 5,697 assets from DexGraspNet~\cite{wang2023dexgraspnet}, each scaled across six geometric sizes ranging from 0.05 m to 0.20 m, are grasped by an Allegro hand using a fingertip style. The baseline's performance directly comes from the paper of Dexonomy.

As shown in Table~\ref{tab: fingertip baseline}, our method consistently outperforms all baseline approaches across every quality metric. Notably, our method \lw{outperforms} Dexonomy by $14\%$ grasp success rate while keeping the zero penetration property. Moreover, \lw{our generated grasps are more diverse}, thanks to the flexibility of hand and object contact points. The running time is discussed in Sec.~\ref{sec:time}.

\begin{figure*}
    \centering
    \includegraphics[width=\linewidth]{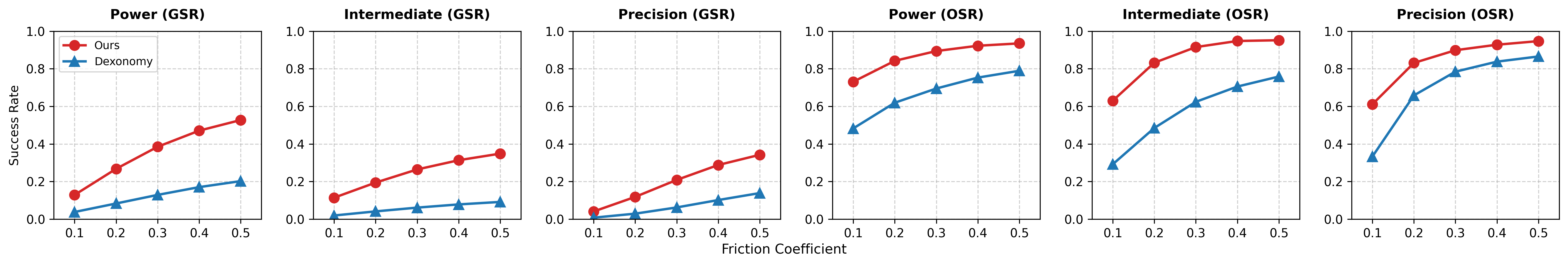}
    \caption{\textbf{Performance on the Hard Benchmark.} Grasp success rate and object success rate across varying tangential friction coefficients. Our method consistently outperforms Dexonomy.}
    \label{fig:harder_6}
\end{figure*}

\begin{table}[t]
    \centering
    \begin{tabular}{l c c c c}
        \toprule
        \textbf{Grasp Type} & \textbf{GSR} (\%) $\uparrow$ & \textbf{OSR} (\%) $\uparrow$ & \textbf{CLN} $\uparrow$ & \textbf{Div} (\%) $\downarrow$ \\
        \midrule
        Power & \textbf{62.8} / 32.9 & \textbf{96.2} / 88.7 & \textbf{12.3} / 9.3 & 26.4 / \textbf{25.2} \\
        Intermediate & \textbf{54.6} / 29.9 & \textbf{96.1} / 89.0 & \textbf{7.4} / 6.4 & 25.6 / \textbf{25.0} \\
        Precision & \textbf{59.2} / 37.5 & \textbf{99.1} / 95.2 & \textbf{5.5} / 4.7 & 25.1 / \textbf{25.0} \\
        \bottomrule
    \end{tabular}%
    \caption{\textbf{Type-Aware Grasp Synthesis Comparison.} Results are presented as (Ours / Dexonomy~\cite{chen2025dexonomy}). We achieve roughly a 100\% relative improvement on the grasp success rate.}
    \label{tab:shadow}
\end{table}

\subsection{Type-Aware Grasp Synthesis}

In this section, we evaluate and compare type-aware grasp synthesis with Dexonomy. Because there are around 30 grasp types, it is time-consuming to test on thousands of objects. Therefore, we randomly select 50 objects from DexGraspNet~\cite{wang2023dexgraspnet} and 50 objects from Objaverse~\cite{deitke2023objaverse}. \lw{This subset is intended as a controlled type-aware evaluation rather than a claim of exhaustive taxonomy-wide validation.} In our early experiments, we find that this subset yields similar performance to the full object set and is thus representative. Following the protocol in Dexonomy~\cite{chen2025dexonomy}, we generate 100 grasp attempts for each object-template pair, and use the same initialization for Dexonomy's refinement stage and our GraspADMM pipeline.

\subsubsection{Quantitative Comparison}
The quantitative statistics for type-aware grasp quality are summarized in Table~\ref{tab:shadow}. Note that the GSR and OSR metrics for Dexonomy are slightly higher than those reported in the original paper due to minor implementation differences and object randomization. Nevertheless, our method significantly outperforms Dexonomy in both GSR and OSR across all three grasp categories, validating the efficacy of our ADMM optimization. Furthermore, the higher Contact Link Numbers (CLN) demonstrate that GraspADMM achieves richer, multi-link contacts, leading to vastly more secure grasps. Our method generates slightly less diverse poses, which is partially due to more samples passing the simulation validation.

\subsubsection{Visual Comparison}
We visually compare our ADMM optimization against the local refinement stage of Dexonomy. As illustrated in Fig.~\ref{fig:vis_compare}, starting from identical initializations, our method yields more visually plausible results for both the Allegro and Shadow hands. While the local refinement stage in Dexonomy fixes contact points on the object and hand links, our method allows adjustments in local frames. Through ADMM optimization, our method encourages lower grasp quality metrics and more contacts, both contribute to more stable grasp poses. \lw{The attached representative convergence traces provide empirical evidence that, despite the nonconvex simulation-based hand update, the optimization stabilizes in representative Allegro and Shadow hand cases. The curve of  the Shadow hand is less smooth compared with the Allegro hand due to higher geometry and contact complexity.}

\begin{figure}[t]
    \centering
    \includegraphics[width=1.\linewidth]{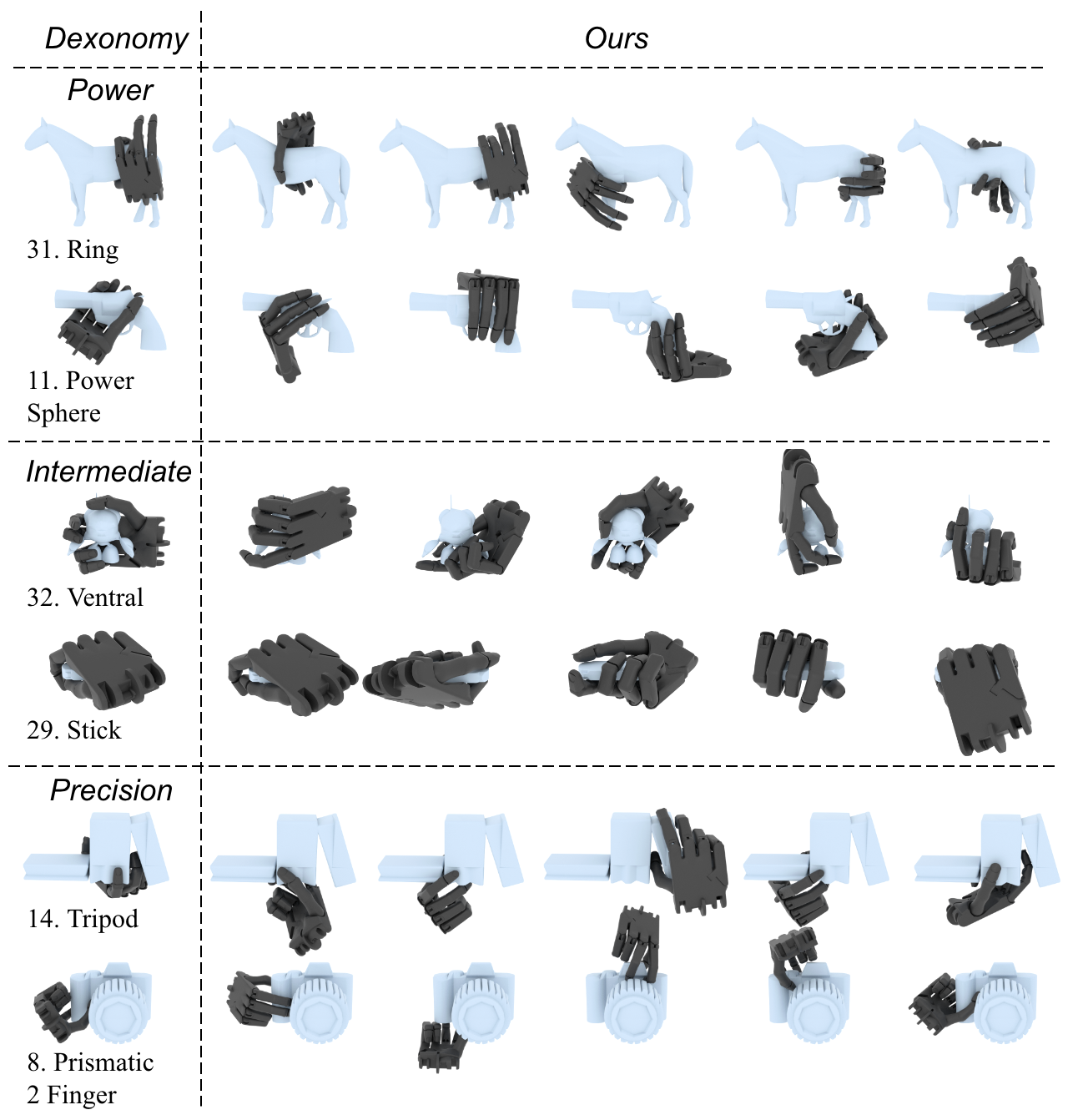}
    \caption{\textbf{Robustness under Extreme Low Friction ($\mu=0.1$).}
    For each displayed object-template pair, Dexonomy produces only one successful grasp among 100 attempts ($\text{GSR} = 1\%$), while GraspADMM produces more successful grasps from the same 100 initializations ($\text{GSR} > 5\%$). The aggregate success rates over the full benchmark are reported in Fig. 4. 
    }
    \label{fig:mu_01}
\end{figure}

\subsection{Robustness for Low-Friction Objects}

The standard simulation validation discussed above assumes tangential and torsional friction coefficients of 0.6 and 0.02, respectively. However, these values often fail to reflect the inherently slippery and unpredictable nature of complex real-world grasping. To rigorously stress-test our approach, we introduce a harder benchmark by eliminating torsional friction entirely (setting it to 0) and incrementally reducing the tangential friction coefficient from 0.5 down to 0.1.

As shown in Fig.~\ref{fig:harder_6}, GraspADMM consistently outperforms Dexonomy in both GSR and OSR across all friction regimes. The relative performance gap actually widens as the environment becomes more challenging, underscoring the necessity of explicitly optimizing dynamic stability.

This advantage is especially pronounced under extreme low-friction conditions ($\mu=0.1$). As visualized in Fig.~\ref{fig:mu_01}, for the selected object-template pairs, Dexonomy rarely produces more than a single grasp pose that survives the simulation validation. In contrast, our decoupled ADMM optimization reliably generates multiple diverse, physically viable grasp poses from the same pool of initial attempts.

\subsection{Ablation Study}

\begin{table}[t]
    \centering
    \begin{tabular}{l c c c c}
        \toprule
         & \textbf{GSR}$\uparrow$ & \textbf{OSR}$\uparrow$ & \textbf{CDC}$\downarrow$ & \textbf{PD}$\downarrow$ \\
        \midrule
        No initialization & 27.2 & 55.1 & 0.31 & \textbf{0.0} \\
        $\rho=0$ & 17.8 & 72.6 & 10.24 & 4.95 \\
        $\rho=10^2$ & 73.5 & 96.7 & 0.21 & \textbf{0.0} \\
        $\rho=10^4$ & 59.3 & 96.2 & 0.29 & \textbf{0.0} \\
        $\rho \rightarrow \infty$ & 62.6 & 94.7 & \textbf{0.04} & \textbf{0.0} \\
        \midrule
        \textbf{Ours ($\rho=10^3$)} & \textbf{74.6} & \textbf{97.2} & 0.17 & \textbf{0.0} \\
        \bottomrule
    \end{tabular}%
    \caption{\textbf{Ablation Study.} A good initialization and a moderate $\rho$ like $10^2$ and $10^3$ is important.}
    \label{tab:ablation}
\end{table}

Table~\ref{tab:ablation} \lw{(captions explained in Sec.~\ref{sec:metric-caption})} shows the impact of different modules and parameter choices on the fingertip grasp synthesis experiment for Allegro hand.

\subsubsection{Initialization}

Our method uses the global alignment stage from Dexonomy~\cite{chen2025dexonomy} as initialization, which first aligns the hand to the object on one contact point, then optimizes the object pose to try to match all contacts. Without this initialization step and no template annotation, the GSR and OSR of our method significantly \lw{drop} to $27.18\%$ and $55.10\%$, showing the importance of initialization. 

\subsubsection{Choice of $\rho$} $\rho$ is the most important hyperparameter in ADMM, as explained in Sec~\ref{sec:admm}. If $\rho=0$, the hand pose is not optimized but directly copied from initialization, \lw{resulting} in low GSR and OSR. If $\rho\rightarrow \infty$, the contact points on the object are fixed, making our \lw{method} perform similarly to Dexonomy~\cite{chen2025dexonomy}. Setting a moderate $\rho$, like $10^2$ and $10^3$, achieves overall high GSR and OSR, but $\rho=10^4$ is too large, causing performance drops. 

\subsubsection{Dual variable annealing}

\begin{figure}
    \centering
    \includegraphics[width=0.8\linewidth]{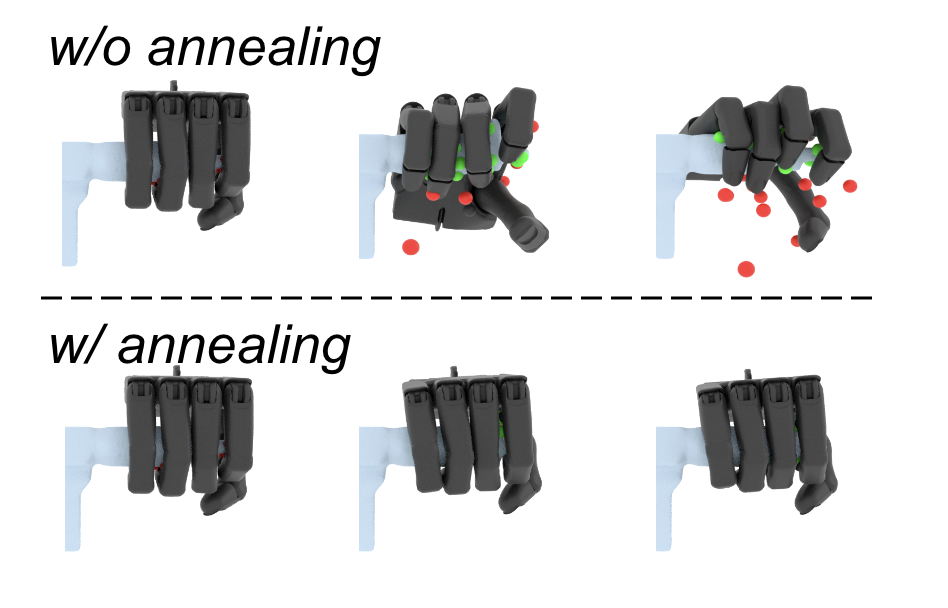}
    \caption{\textbf{Effect of Dual Variable Annealing.} Green: target object contacts $\mathbf{p}^o$. Red: dual-shifted targets $\mathbf{p}^o - \lambda$. 
    }
    \label{fig:annealing}
\end{figure}

We visualize the importance of dual variable annealing in Fig.~\ref{fig:annealing}. The green points represent the object contact points $\mathbf{p}^o$, the red points represent $\mathbf{p}^o - \lambda$. Without annealing, due to the nonzero gap between $\mathbf{p}^h$ and $\mathbf{p}^o$, $\lambda$ accumulates rapidly across iterations, causing $\mathbf{p}^o - \lambda$ to drastically deviate from $\mathbf{p}^o$. This makes the hand become unstable in the second stage of ADMM. After applying annealing, $\lambda$ is controlled under a certain threshold, \lw{allowing} the hand to stably squeeze as tight as possible.

\subsection{Time Analysis}
\label{sec:time}
In terms of optimization speed, both our method and Dexonomy utilize a hybrid CPU (for MuJoCo) and GPU pipeline, whereas the other baselines rely primarily on GPU optimization. On a single NVIDIA RTX 4090 GPU, DexGraspNet~\cite{wang2023dexgraspnet}, FRoGGeR~\cite{li2023frogger}, and SpringGrasp~\cite{chen2024springgrasp} average fewer than 3 samples per second (s/s). BODex~\cite{chen2024bodex} reaches approximately 40 s/s. Utilizing a single AMD EPYC 7543 2.8GHz CPU, Dexonomy processes around 24 s/s. Our method requires slightly more computation time due to the iterative ADMM optimization, processing approximately 19 s/s. However, while BODex boasts the highest throughput, it is strictly limited to fingertip-style grasping, whereas our framework inherently supports the full taxonomy of diverse grasp types. Furthermore, because our method achieves a significantly higher GSR than Dexonomy, the actual yield of successful, usable grasp poses generated per second is highly comparable \lw{(14.5 s/s for Dexonomy and 14.2 s/s for GraspADMM)}. Future extensions, such as scaling CPU resources or transitioning to GPU-native simulators like NVIDIA Newton~\cite{newton}, could further accelerate our pipeline.
\section{Conclusion \& Limitation}

In this work, we presented GraspADMM, a novel dexterous grasp synthesis framework that improves the kinematic feasibility and dynamic stability. By formulating the refinement stage via the Alternating Direction Method of Multipliers (ADMM), our method successfully decouples the object contact points from hand contact points, enabling the efficient minimization of grasp metrics while enforcing penetration-free kinematic constraints. 
Extensive experiments demonstrate that GraspADMM achieves state-of-the-art success rates across diverse grasp types and object geometries, \lw{remaining} robust even under extreme low-friction conditions (e.g., $\mu=0.1$).

Nevertheless, our work faces several limitations. \lw{First, all experiments are currently conducted in simulation. Although MuJoCo enables controlled evaluation of penetration-free grasp stability, real-robot and cross-simulator validation are still needed to assess sim-to-real transfer and to verify that the reported gains are not specific to a single physics engine. Second, our type-aware evaluation uses a controlled subset of objects and grasp categories; therefore, it should be viewed as representative evidence rather than exhaustive validation over the full taxonomy. Third, GraspADMM is currently an offline grasp-candidate refinement method. The optimized grasp is generated before execution and passed to a downstream controller or planner, while online closed-loop adaptation would require modeling object motion, slip, and sensing feedback. Finally, our CPU-based MuJoCo implementation limits raw throughput, though the independent optimization of grasp candidates makes GPU-native simulation and larger-scale parallelization natural future directions. Future work will also explore large-scale dataset generation, trajectory-level execution, and cluttered scenes.}

\bibliographystyle{IEEEtran}
\bibliography{ref}


\end{document}